\documentclass[paper=a4]{article}
\pdfoutput=1
\usepackage{PRIMEarxiv}

\usepackage[utf8]{inputenc}
\usepackage[T1]{fontenc}
\usepackage{hyperref}
\usepackage{amsfonts}
\usepackage{fancyhdr}       
\usepackage{graphicx}       
\usepackage{multirow}
\usepackage{float}
\usepackage{caption}
\usepackage{longtable}
\usepackage{amsmath}

\usepackage[
  backend=biber,
  style=ieee,
]{biblatex}
\graphicspath{{.}}

\title{The Effectiveness of a Dynamic Loss Function in Neural Network Based Automated Essay Scoring}

\author{
  Oscar Morris \\
  \texttt{twocap06@gmail.com} \\
}

\begin{document}
\maketitle

\begin{abstract}
Neural networks and in particular the attention mechanism have brought significant advances to the field of Automated Essay Scoring. Many of these systems use a regression-based model which may be prone to underfitting when the model only predicts the mean of the training data. In this paper, we present a dynamic loss function that creates an incentive for the model to predict with the correct distribution, as well as predicting the correct values. Our loss function achieves this goal without sacrificing any performance achieving a Quadratic Weighted Kappa score of 0.752 on the Automated Student Assessment Prize Automated Essay Scoring dataset.
\end{abstract}

\section{Introduction}

Automated Essay Scoring (AES) is the task of assigning a score to free-form text (throughout this paper essay will be defined loosely to include short answers) using a computational system. The goal of AES is to mimic human scoring as closely as possible. The development of the Transformer in \cite{bahdanauNeuralMachineTranslation2016} has significantly improved the performance of Natural Language Processing (NLP) models to a point where it is achievable to use a purely neural approach to AES \cite{taghipourNeuralApproachAutomated2016,ludwigAutomatedEssayScoring2021}. This has created the possibility for many task-agnostic architectures and pre-training approaches which then allows for greater flexibility in the implementation of these models. This also makes the cutting-edge performance of these NLP models available for simple implementation in real world situations.

Transformer models such as those in \cite{vaswaniAttentionAllYou2017,devlinBERTPreTrainingDeep2019,beltagyLongformerLongDocumentTransformer2020} and many more have the significant disadvantage that they require a very large training dataset and high training time to achieve decent performance. This makes training these models from scratch almost unachievable for most tasks (which don't have very large datasets available). A good solution to this is pre-training where the model is first trained by the creator on a task-agnostic dataset and, if necessary, the model can then be fine-tuned to the downstream task that it will be applied to. After fine-tuning these models can achieve the performance increase of a transformer model with minimal training time and a small training dataset.

All sequence to sequence transformer models use an encoder-decoder architecture where the input sequence is fed into an encoder. The encoder then outputs a vector or matrix representation of the input sequence known as the `hidden vector' or `hidden matrix'. This can then be inputted into a decoder which converts this hidden data back into an output sequence. Since the encoder and decoder are independent networks, the decoder can be replaced with a classification or regression head, taking the hidden data as its input and outputting multiple or a single neuron value. This significantly expands the tasks transformers can be applied to.

The encoders and decoders do not have to use a transformer architecture, they can be replaced by older Recurrent Neural Networks (RNNs). Even though RNNs often perform worse than transformer models, on small datasets they do often perform similarly or better than transformer models (even if the transformer is pre-trained).

For the task of AES there has been discussion on whether a classification approach is better than a regression approach (and vice versa) \cite{johanberggrenRegressionClassificationAutomated2019}. It was found that a regression approach performs better for their dataset, however, if a pre-training approach is used for an AES system classification has the significant disadvantage that the number of marks given cannot be changed between the pre-training step and the fine-tuning step, whereas with a regression approach the number of marks is irrelevant which may significantly improve the size of the dataset that can be used, or even if a pre-training approach is reasonable.

The distribution of scores given to a set of essays is naturally very unbalanced as the aim for human markers is often to obtain a normal distribution of the scores. This means that there are few answers that achieve the very worst scores and very few that achieve the highest scores. Unbalanced datasets can significantly reduce the performance of a classification model. Unbalanced datasets can pose an issue for a regression model in that it can be easier for the model to only predict the mean of the training data, this is a significant problem for AES as it is very important that each sample is given as accurate a score as possible. The loss function proposed in this paper aims to solve this problem.

\section{Dataset}

The dataset used to train this model was introduced in the Kaggle Automated Student Assessment Prize (ASAP) in 2012.\footnote{\url{https://kaggle.com/c/asap-aes}} For the ASAP competition two datasets were introduced, one containing essays (approx. 150-650 words) and the other containing short answers (\textless{} 100 words). This model is trained only on the essay dataset. Table \ref{tbl:asap-aes} shows the details of the dataset.

\begin{table}[H]
\captionsetup{justification=centering}
\centering
\caption{Details of ASAP AES dataset \label{tbl:asap-aes}}
\begin{tabular}{|l|l|l|}
\hline
Prompt & No. of Samples & Score Range \\ \hline
1      & 1,783          & 2-12        \\
2      & 1,800          & 1-6         \\
3      & 1,726          & 0-3         \\
4      & 1,772          & 0-3         \\
5      & 1,805          & 0-4         \\
6      & 1,800          & 0-4         \\
7      & 1,569          & 0-30        \\
8      & 723            & 0-60        \\ \hline
\end{tabular}
\end{table}

The model was trained on each prompt individually. First the dataset is shuffled with 90\% of the data used for training and the other 10\% used for evaluation.

\section{Model}

\subsection{Architecture}

The model used in this paper is a Long Short-Term Memory (LSTM) \cite{hochreiterLongShortTermMemory1997} encoder using an attention mechanism \cite{bahdanauNeuralMachineTranslation2016}. The regression head used for this model is a single fully connected layer where the hidden data is inputted and a single value is outputted.

To tokenise the input sequence the same tokeniser used in the BERT model \cite{devlinBERTPreTrainingDeep2019} is used. This choice was made because the BERT tokeniser tokenises on word-parts rather than whole words. This allows the model to be more confident in the meaning of words that are not in its original training data, which may occur when using technical language in an essay.

\begin{figure}[H]
\centering
\includegraphics[height=8cm]{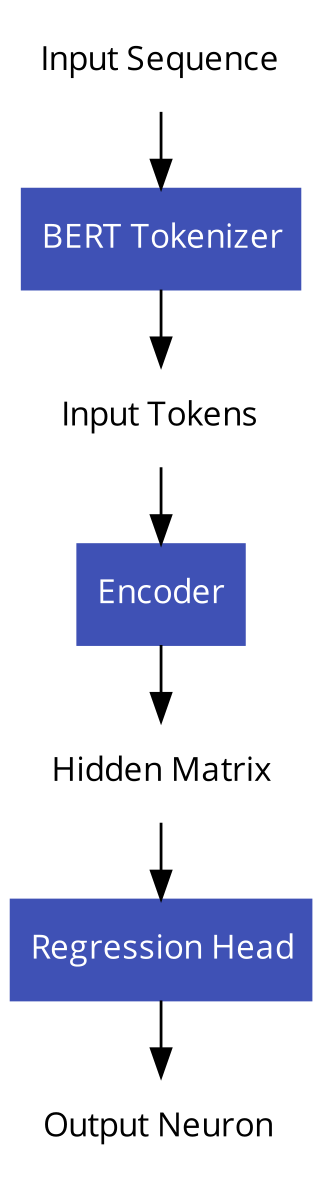}
\caption{Model architecture}
\label{fig:arch}
\end{figure}

\begin{figure}[H]
\centering
\includegraphics[height=6cm]{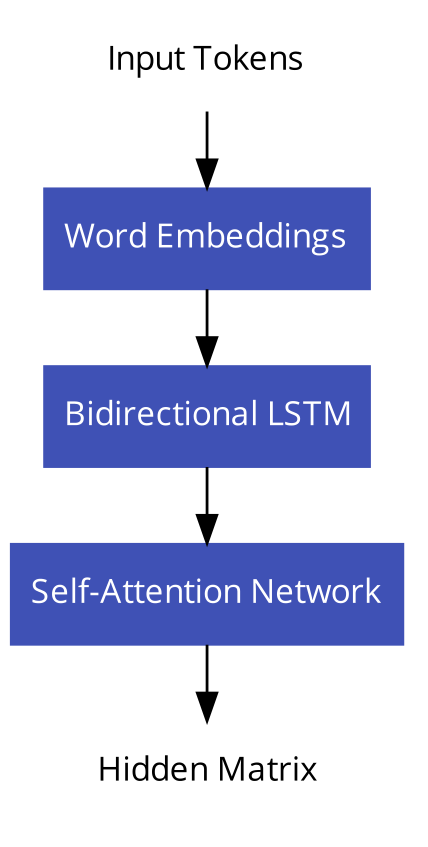}
\caption{Encoder architecture}
\label{fig:encoder}
\end{figure}

Fig. \ref{fig:arch} shows the flow of data through the model and Fig. \ref{fig:encoder} takes a closer look at the architecture of the encoder.\footnote{Images created with \href{https://code2flow.com}{code2flow}}

\subsection{Metrics}

In the ASAP competition the metric used was the Quadratic Weighted Kappa (QWK) defined in \cite{taghipourNeuralApproachAutomated2016}.

The model was also evaluated on Mean Squared Error (MSE), defined in Eq. \ref{eq:mse}, Mean Absolute Error (MAE), defined in Eq. \ref{eq:mae} and the coefficint of determination \(r^2\), defined in Eq. \ref{eq:r2}.

\begin{equation} \text{MSE}(\mathbf{x}, \mathbf{y}) = \frac{1}{N} \sum^{N}_{i=0} (y_i-x_i)^2 \label{eq:mse}\end{equation}

\begin{equation} \text{MAE}(\mathbf{x}, \mathbf{y}) = \frac{1}{N} \sum^N_{i=0} |y_i-x_i| \label{eq:mae}\end{equation}

\begin{equation} r^2 = 1-\frac{\sum^N_{i=0} (y_i-x_i)^2}{\sum^N_{i=0} (y_i-\bar{y})} \label{eq:r2}\end{equation}

where \(\mathbf{x}\) are the true values, \(\mathbf{y}\) are the predicted values, \(N\) is the number of samples and \(\bar{y}\) is the mean predicted value.

\section{The Dynamic Loss Function}

A loss function is the function that is minimized during the training of the network. The aim of a loss function is to determine the difference between what the models outputs are and what they should be. A dynamic loss function is when the loss function is changed throughout the training process. The main benefit of a dynamic loss function is the ability to adjust the goals of the model at different times during the training process.

A common problem with regression models is that the model tends towards predicting the mean of the training dataset with very little variation. This can occur when the dataset is unbalanced (as is often the case with real world datasets), however, in prior testing it did occur on the ASAP AES dataset.

A solution to this problem is to adjust the loss function to provide an incentive to predicting batches with the correct sample standard deviation. Do do this, a loss function can be defined as the error in the standard deviation of a certain batch:

\begin{equation} \text{STDE}(\mathbf{x}, \mathbf{y}) = |\sigma(\mathbf{y})-\sigma(\mathbf{x})| \label{eq:stderror}\end{equation}

where \(\sigma\) is the function calculating the sample standard deviation.

Multiple loss functions can be combined using a weighted sum, and some constant \(p\):

\begin{equation} L_T = pL_1 + (1-p)L_2 \label{eq:combine-loss}\end{equation}

where \(L_T\) is the total loss and \(L_1\) and \(L_2\) are two loss functions.

Therefore, using \(\text{STDE}\) as \(L_1\) and \(\text{MSE}\) as \(L_2\) a loss function is defined that provides an incentive to predict using the correct standard deviation and with minimal error.

\begin{figure}[H]
\centering
\includegraphics[height=6cm]{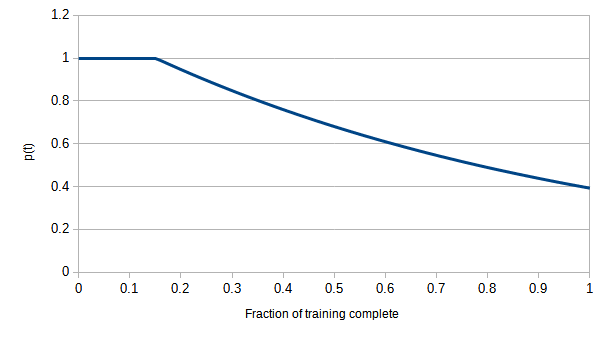}
\caption{Value of $p$ against fraction of training complete using $a=1$, $b=0.15$ and $c=1.1$}
\label{fig:p}
\end{figure}

Using a constant value of \(p\) did help to prevent this form of underfitting, however, the model was still showing significant underfitting and because of the reduced importance of an error metric the model's performance was significantly reduced. In an attempt to solve both of these problems, the value of \(p\) could be decayed over time using an exponential decay function (shown in Eq. \ref{eq:decay}). When the decay was started as soon as the training began it was found that the model would still show signs of underfitting, this could easily be solved by holding \(p\) constant for the first portion of training and then decaying its value throughout the rest of the training process. The full definition of \(p\) as a function of training step or epoch is shown in Eq. \ref{eq:min-decay}.

\begin{equation} p(t) = a\cdot \exp\left(-\frac{t}{T}\right) \label{eq:decay}\end{equation}

\begin{equation} p(t) = \min\left(a, a \cdot \exp\left(-c\left(\frac{t}{T}-b\right)\right)\right) \label{eq:min-decay}\end{equation}

where \(t\) is the current training step or current epoch, \(T\) is the total number of training step or total number of training epochs and \(a\), \(b\) and \(c\) are constants. Fig. \ref{fig:p} shows an example plot of \(p\) against fraction of training complete.

This method achieved the highest performance without sacrificing either metric.

\section{Results}

The winner of the ASAP competition is the Enhanced AI Scoring Engine (EASE) \cite{EASE2022}. EASE is a feature-extraction based system where the features are then used to build a regression model using either Support Vector Regression (SVR) or Bayesian Linear Ridge Regression (BLRR). We compare our model to both EASE systems and two fully neural models using LSTM, one using the Mean over Time (MoT) and the other using an attention mechanism as described in \cite{taghipourNeuralApproachAutomated2016}. Unfortunately, Taghipour and Ng did not release detailed data for their attention model, only the mean QWK score it achieved across all prompts.

\begin{table}[H]
\captionsetup{justification=centering}
\centering
\caption{Our model compared to EASE and Taghipour and Ng's LSTM model with Mean over Time and Attention \label{tbl:results}}
\begin{tabular}{|l|ccccccccc|}
\hline
\multirow{2}{*}{System} & \multicolumn{9}{c|}{Prompt}                                                                  \\ \cline{2-10} 
                        & 1              & 2              & 3              & 4              & 5              & 6              & 7     & \multicolumn{1}{c|}{8}     & Avg QWK \\ \hline
Ours                    & \textbf{0.841} & 0.663          & 0.672          & 0.773          & 0.807          & 0.743          & \textbf{0.839} & \multicolumn{1}{c|}{\textbf{0.679}} & \textbf{0.752}   \\
EASE (SVR)              & 0.781          & 0.621          & 0.630          & 0.749          & 0.782          & 0.771          & 0.727 & \multicolumn{1}{c|}{0.534} & 0.699   \\
EASE (BLRR)             & 0.761          & 0.606          & 0.621          & 0.742          & 0.784          & 0.775          & 0.730 & \multicolumn{1}{c|}{0.617} & 0.705   \\
Taghipour and Ng MoT    & 0.775          & \textbf{0.687} & \textbf{0.683} & \textbf{0.795} & \textbf{0.818} & \textbf{0.813} & 0.805 & \multicolumn{1}{c|}{0.594} & 0.746   \\
Taghipour and Ng Attn.  &                &                &                &                &                &                &       & \multicolumn{1}{c|}{}      & 0.731   \\ \hline
Human graders           & 0.721          & 0.812          & 0.769          & 0.851          & 0.753          & 0.776          & 0.720 & \multicolumn{1}{c|}{0.627} & 0.754   \\ \hline
\end{tabular}
\end{table}

As can be seen in Table \ref{tbl:results}, Our system outperforms both EASE models by a significant margin in all prompts and performs approximately equivalently to the models proposed by Taghipour and Ng. The model improves slightly the score achieved by both models proposed by Taghipour and Ng. This may seem surprising as it may be assumed that since the error-based loss is 'less important' to the model, the model's performance would decrease. However, these results show this is not the case. The QWK scores of these models are all close to the difference between human graders with our approach being effectively equivalent.

\begin{figure}[H]
\captionsetup{justification=centering}
\centering
\includegraphics[height=6cm]{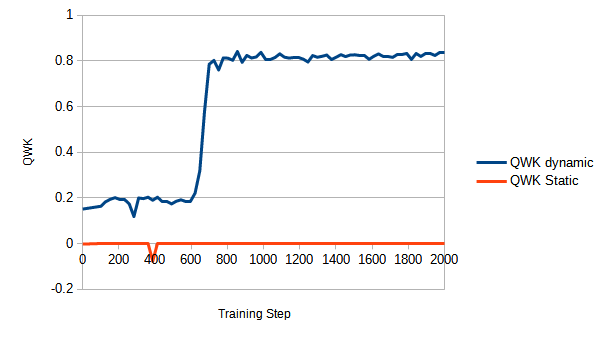}
\caption{QWK on Prompt 1. Blue: Our Dynamic Loss Function, Orange: MSE loss}
\label{fig:qwk}
\end{figure}

\begin{figure}[H]
\captionsetup{justification=centering}
\centering
\includegraphics[height=6cm]{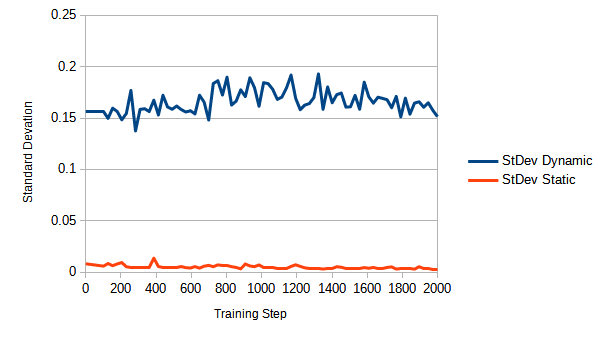}
\caption{Standard deviation on Prompt 1, the standard deviation of the evaluation split is 0.17. Blue: Our Dynamic Loss Function, Orange: MSE loss}
\label{fig:stdev}
\end{figure}

The main goal of these experiments was to create a loss function that prevents underfitting on regression tasks. To show that this has been achieved, the model was trained twice on prompt 1 and the QWK and standard deviation of its predictions on the evaluation split were measured after every epoch. The results of this is shown in Figures \ref{fig:qwk} and \ref{fig:stdev}.

These figures show that the performance of our loss function is a significant improvement over only using MSE which causes severe underfitting, shown by the extremely low standard deviation compared to the actual standard deviation of the dataset. The QWK score of the model trained only using MSE loss is 0 which implies that any agreement between the model and the human grader is by chance.

\section{Conclusion}

The dynamic loss function proposed in this paper significantly improves upon other loss functions in reducing underfitting in regression tasks. Our loss function has also shown that it does not sacrifice the performance of the model, it even improves on the performance achieved by other approaches. Our model makes use of an attention mechanism that allows the model to weight the importance of different tokens in the input sequence. This eliminates the need for handcrafted features which can be inaccurate and time-consuming to create. The large disadvantage of using neural networks for AES is the increased compute power required. However, using an LSTM-based model instead of a transformer-based model significantly improves this.

\printbibliography

\end{document}